\documentclass[twoside]{article}

\usepackage[accepted]{aistats2026}
\usepackage{soul}
\usepackage[round]{natbib}

\usepackage{algorithm}
\usepackage{algorithmic}

\usepackage[utf8]{inputenc} 
\usepackage[T1]{fontenc}    
\usepackage{url}            
\usepackage{booktabs}       
\usepackage{amsfonts}       
\usepackage{nicefrac}       
\usepackage{microtype}      
\usepackage{xcolor}         
\usepackage{multirow}
\usepackage{hyperref}
\usepackage{amsmath, amssymb, amsthm}
\usepackage{graphicx}
\usepackage{wrapfig}

\newcommand{\minisection}[1]{\noindent{\textbf{#1}}}

\newtheorem{definition}{Definition}
\newtheorem{property}{Property}

\begin{document}

%

%

\twocolumn[

\aistatstitle{Data Distribution Valuation Using Generalized Bayesian Inference}

\aistatsauthor{Cuong N. Nguyen \And Cuong V. Nguyen }

\aistatsaddress{Department of Mathematical Sciences \\ Durham University \And Department of Mathematical Sciences \\ Durham University} ]

\begin{abstract}
We investigate the data distribution valuation problem, which aims to quantify the values of data distributions from their samples. This is a recently proposed problem that is related to but different from classical data valuation and can be applied to various applications. For this problem, we develop a novel framework called \emph{Generalized Bayes Valuation} that utilizes generalized Bayesian inference with a loss constructed from transferability measures. This framework allows us to solve, in a unified way, seemingly unrelated practical problems, such as annotator evaluation and data augmentation. Using the Bayesian principles, we further improve and enhance the applicability of our framework by extending it to the continuous data stream setting. Our experiment results confirm the effectiveness and efficiency of our framework in different real-world scenarios. Our code is available at: \url{https://github.com/CuongNN218/GBV}.
\end{abstract}

\section{INTRODUCTION}
\label{sec:intro}

Data distribution valuation is a recently proposed problem that aims to estimate and compare values of data distributions from their finite samples~\citep{amiri2022fundamentals, xu2024data}. This problem is related to but distinguished from traditional data valuation, which focuses on estimating the values of single data points~\citep{kwon2021beta, just2023lava, kessler2024sava}. Data distribution valuation is often required in many real-world scenarios, such as when data buyers evaluate the quality of data from different vendors before making a purchase~\citep{xu2024data}.

The main approaches for data distribution valuation mainly estimated the differences between seller's sample data and buyer's reference data~\citep{amiri2022fundamentals, xu2024data}, but they often made restrictive assumptions to assess the distributions. For instance, \citet{amiri2022fundamentals} required a central broker with full data access, imposing potential privacy risks to the buyer and sellers. Subsequent work by~\citet{xu2024data} removed the broker and utilized maximum mean discrepancy (MMD) to directly estimate the distance between the sample and reference data.

In this paper, we propose a novel and general framework for data distribution valuation that does not require the above assumptions. Our framework, called \emph{Generalized Bayes Valuation} (GBV), utilizes generalized Bayesian inference~\citep{bissiri2016general, matsubara2024generalized} to construct a posterior over the data sources (i.e., the data distributions to be evaluated) and return this posterior as the valuation. In our GBV framework, the classical negative loglikelihood is replaced by a general loss function constructed from transferability measures in transfer learning~\citep{nguyen2020leep, nguyen2023simple, you2021logme, gholami2023etran}, with MMD~\citep{xu2024data} being a special case. Due to its simplicity and the computational efficiency of transferability measures, GBV is highly scalable and data efficient, making it attractive for large-scale applications with many data sources. As specific instances of the framework, we demonstrate how GBV can be applied to annotator evaluation~\citep{xu2024data} and to data augmentation~\citep{yang2024entaugment}, a surprising application that may initially appear unrelated.

Additionally, by leveraging the inherent properties of Bayesian inference in continuous data stream settings~\citep{nguyen2024lifelong}, we extend our framework to develop \emph{Continual Generalized Bayes Valuation} (CGBV), a solution for the dynamic scenario where data from each source is provided sequentially. This new scenario is common in real-world applications where data buyers continuously acquire data over several episodes and need to re-evaluate the sellers' data quality without access to data from past episodes.

In summary, our paper makes the following contributions: (1) We develop the novel GBV framework that employs generalized Bayesian inference with a transferability loss for data distribution valuation; (2) We show how GBV can be applied to the annotator evaluation and data augmentation problems; (3) We extend GBV to CGBV, a solution for data distribution valuation in the continuous data stream setting; and (4) We empirically show the effectiveness of our methods on different datasets for annotator evaluation, data augmentation, and continual annotator evaluation.

\section{RELATED WORK}
\vspace{-2mm}

Our work is related to data distribution valuation, generalized Bayesian inference, and transferability estimation. We discuss the most relevant work below.

\minisection{Data Distribution Valuation.} Current work on data distribution valuation mainly estimated the differences between the seller's sample data and buyer's reference data. For example, \citet{amiri2022fundamentals} estimated these differences by calculating the variance of the seller's data projected onto the principal components of the buyer's reference data. On the other hand, \citet{xu2024data} used the negated mean discrepancy to empirically measure the distance between the seller and buyer data distributions. These studies, however, often relied on some restrictive assumptions. \citet{amiri2022fundamentals} required sharing the buyer's principal components to the seller, imposing privacy risks that necessitate a trusted third-party broker. \citet{xu2024data} removed the broker but assumed data followed a Huber model to facilitate theoretical analysis. In contrast, our work relaxes these assumptions, generalizes the approach of \citet{xu2024data} through the use of transferability measures, and considers novel applications of data distribution valuation, including a new setting for continual data distribution valuation.

\minisection{Generalized Bayesian Inference.} Generalized Bayesian inference offers a robust and flexible framework to update prior beliefs, particularly in scenarios where the likelihood is misspecified~\citep{matsubara2024generalized}, intractable~\citep{pacchiardi2024generalized}, or even when the probabilistic model is not explicitly defined \citep{bissiri2016general}. Its core principle is to substitute the negative loglikelihood with a general loss function. For example, \cite{holmes2017assigning} addressed the misspecification by down-weighting the likelihood by a factor $\alpha$, which is a special case of the loss-based method~\citep{bissiri2016general, loaiza2021focused, pacchiardi2024generalized}. In this direction, various loss functions have been proposed to measure the divergence between the statistical model and empirical data distribution, including the kernel Stein discrepancy~\citep{matsubara2022robust}, maximum mean discrepancy~\citep{pacchiardi2024generalized},  and Fisher divergence~\citep{matsubara2024generalized}. Due to its flexibility, generalized Bayesian inference has been applied to different areas, including uncertainty quantification~\citep{charpentier2020posterior} and model inversion attacks~\citep{wang2021variational}. In this work, we will show a new application of this framework, namely data distribution valuation.

\minisection{Transferability Estimation.} Transferability measures are a practical tool in transfer learning to assess how effectively knowledge from a source dataset or pre-trained model transfers to a target dataset. Some methods leverage label distributions to quantify source–target correlations~\citep{tran2019transferability, nguyen2020leep, nguyen2022generalization}, although these often rely on unrealistic shared-input assumptions~\citep{tran2019transferability} or suffer from overfitting~\citep{nguyen2020leep, nguyen2022generalization}. To overcome these issues, subsequent work has focused on feature-space analyses, including domain distance~\citep{tan2021otce}, feature-label alignment~\citep{you2021logme}, and intra/inter-class distance metrics~\citep{bao2019information, pandy2022transferability, ibrahim2022newer}. Notably, the potential energy of source models on target data has shown strong predictive power for transfer performance~\citep{gholami2023etran}. These measures can support applications in model selection~\citep{you2021logme, li2021ranking} and model ensembling~\citep{agostinelli2022ensemble, bachu2023building}. However, their application in data distribution valuation was largely unexplored and our work is the first to integrate them with generalized Bayesian inference for this purpose.

\section{PROBLEM SETTING}
\label{sec:setting}
\vspace{-2mm}

We study the \emph{data distribution valuation} problem proposed by~\cite{xu2024data} for classification. In this problem, we have a set $\mathcal{S}$ of data sources (also called data vendors), where each element $s \in \mathcal{S}$ corresponds to an unknown data distribution $\mathbb{P}_s (X, Y)$. In general, the set $\mathcal{S}$ could be countably infinite  or continuous. For each data source $s \in \mathcal{S}$, we are given a sample set $\mathcal{D}_s = \{ (x_{s,i}, y_{s,i})\}_{i=1}^{N_s}$, where $(x_{s,i}, y_{s,i}) \sim \mathbb P_s$, which is representative of the training data that we can get from this source. Assume a model trainer (or data buyer) would like to evaluate the suitability (or valuation) of the data sources for a target task with unknown test data distribution $\mathbb{P}^*(X, Y)$. Since $\mathbb{P}^*$ is unknown, the trainer can only access a validation (reference) set ${ \mathcal{D}^* = \{ (x^*_i, y^*_i)\}_{i=1}^{N^*} }$, where $(x^*_i, y^*_i) \sim \mathbb{P}^*$. The set $\mathcal{D}^*$ will be used as the validation set for valuation since it has the same distribution $\mathbb{P}^*$ as the actual test set.

The valuation of the data sources can be represented by a distribution $P(s)$ constructed from $\mathcal{D}^*$ and all $\mathcal{D}_s$'s. This distribution will be utilized by the model trainer during training to achieve a good accuracy on the test set. A common way to use $P(s)$ is to minimize the following weighted cross-entropy loss on a given training set $\{ (x_{s,i}, y_{s,i}) \}_{{s \in \mathcal{S}}, 1 \le i \le n_s}$: \\[-2mm]
\begin{equation}
\ell(\theta) = \mathbb{E}_{s \sim P(s)} \Big[ \frac{1}{n_s} \textstyle{\sum_{1 \le i \le n_s}} \ell (\theta; x_{s,i}, y_{s,i}) \Big],
\label{eq:weighted_loss}
\vspace{-0.5mm}
\end{equation}
where $\theta$ is the model parameter vector and (with a slight abuse of notation) $\ell (\theta; x, y)$ is the cross-entropy loss for the data point $(x, y)$. In this case, the data distribution valuation problem can be stated as follows.

\minisection{Data distribution valuation.} \emph{Given the sets $\mathcal{D}^*$ and $\mathcal{D}_s$ for all $s \in \mathcal{S}$, find a distribution $P(s)$ of the data sources such that if we train a model $m_{\theta^*}$ by minimizing~\eqref{eq:weighted_loss}, then $m_{\theta^*}$ minimizes the expected error $\mathbb{E}_{(x,y) \sim \mathbb{P}^*} [ y \neq m_{\theta^*}(x)]$ on the target task.}

We note a few important differences between this problem setting and that of~\cite{xu2024data}:

$\bullet$ While their goal focuses on deriving a distribution valuation function $\Upsilon$ to correctly measure the negative distance between a source and target distributions, our objective is more target-oriented. That is, our data source distribution $P(s)$ focuses on facilitating the training of the target model, even with noisy or distribution-drifted training sets.

$\bullet$ They define ${ \Upsilon(\mathbb{P}_s) = -d(\mathbb{P}_s, \mathbb{P}^*) }$ based on a distance $d$ between two distributions and constructs a solution $\nu(\mathcal{D}_s)$ that can approximately maintain the order of $\Upsilon$, i.e., $\Upsilon(\mathbb{P}_s) \gtrsim \Upsilon(\mathbb{P}_{s'})$ if $\nu(\mathcal{D}_s) > \nu(\mathcal{D}_{s'})$. This restricts $\nu$ to be dependent on $d$ and may potentially limit its generalizability. Our setting, in contrast, does not assume the existence of $\Upsilon$ and $d$. The only requirement for our solution $P(s)$ is a good target test error $\mathbb{E}_{(x,y) \sim \mathbb{P}^*} [ y \neq m_{\theta^*}(x)]$, which is more natural and realistic for evaluating the quality of $P(s)$.

An application of data distribution valuation is annotator evaluation~\citep{xu2024data}. In this application, each data source $s$ is an annotator who gives labels to training data. However, the quality of labels from different annotators may vary, with each annotator having some label noise. Thus, the model trainer would like to find a valuation distribution $P(s)$ based on their gold standard validation set $\mathcal{D}^*$ and the sample sets $\mathcal{D}_s$ collected from the annotators. This distribution will be used with the loss~\eqref{eq:weighted_loss} to improve the accuracy of the target model.

We will later show that data distribution valuation can also be applied to derive a solution for another application: data augmentation~\citep{cubuk2019autoaugment, cubuk2020randaugment, muller2021trivialaugment}. This is a surprising and interesting finding since annotator evaluation and data augmentation seem unrelated on the surface. But first, we will present in the next section, a unified approach for the general data distribution valuation problem.

\section{GENERALIZED BAYESIAN DATA DISTRIBUTION VALUATION}
\label{sec:gbv}

To construct $P(s)$ for the data distribution valuation problem above, we propose a new method based on generalized Bayesian inference~\citep{bissiri2016general}. The main idea of our method is to place a prior distribution $p(s)$ over $\mathcal{S}$ and update this prior belief to a posterior $p(s \,|\, \{\mathcal{D}_{s'}\}_{s' \in \mathcal{S}})$ through a special loss function $L(s, \mathcal{D}_s, \mathcal{D}^*)$. This is generalized Bayesian inference because it uses a loss function to connect information between $s$ and the data $\mathcal{D}_s$ instead of a traditional likelihood function~\citep{bissiri2016general}. We note that the loss $L$ is different from the training loss such as~\eqref{eq:weighted_loss}, and thus, in this paper, we shall refer to $L$ as the generalized Bayesian loss.

Our method, called \emph{Generalized Bayes Valuation} (GBV), is as follows. Consider a prior distribution $p(s)$ over $s \in \mathcal{S}$. If $\mathcal{S}$ is finite with $|\mathcal{S}| = M$ (e.g., there are $M$ label annotators), then $p(s)$ is a probability mass function with support $\{ 1, 2 \ldots, M \}$. If $\mathcal{S}$ is uncountable (e.g., $\mathcal{S} \subseteq \mathbb{R}$), then $p(s)$ is a density function on $\mathcal{S}$. More complicated priors can also be constructed from $\mathcal{D}^*$, but we do not explore them here for simplicity.

Assume we have a generalized Bayesian loss function ${ L(s, \mathcal{D}_s, \mathcal{D}^*) \in \mathbb{R} }$. Applying the inference method in~\citet{bissiri2016general}, we can replace the likelihood function $p(\{\mathcal{D}_{s'}\}_{s' \in \mathcal{S}} \,|\, s)$ in Bayes' rule by $e^{-L(s, \mathcal{D}_s, \mathcal{D}^*)}$. Note that by using this likelihood, we implicitly assume that $p(\{\mathcal{D}_{s'}\}_{s' \in \mathcal{S}} \,|\, s)$ only depends on $\mathcal{D}_s$ and does not depend on $\{\mathcal{D}_{s'}\}_{s' \neq s}$. Thus, we will write $p(\{\mathcal{D}_s\} \,|\, s )$ and $p(s \,|\, \{ \mathcal{D}_s \})$ as short-handed notations for $p(\{\mathcal{D}_{s'}\}_{s' \in \mathcal{S}} \,|\, s  )$ and $p(s \,|\, \{\mathcal{D}_{s'} \}_{s' \in \mathcal{S}} )$, respectively. With the likelihood above, we can obtain the posterior $p(s \,|\, \{ \mathcal{D}_s \})$ by:
\begin{align}
    p(s | \{ \mathcal{D}_s \}) &\,\propto\, p(s) \, p(\{ \mathcal{D}_s \} | s) \,\propto\, p(s) \, e^{-L(s, \mathcal{D}_s, \mathcal{D}^*)}.
    \label{eq:gbv}
\end{align}
We assume here that $\mathcal{D}^*$ is constant and given to the algorithm. Thus, the prior $p(s)$, the likelihood $p(\{\mathcal{D}_s\} \,|\, s)$, and the loss $L$ may depend on $\mathcal{D}^*$. From Eq.~\eqref{eq:gbv}, if we can construct and compute an appropriate loss $L$, then we can compute the posterior $p(s \,|\, \{\mathcal{D}_s\})$ and use it as the solution for the data distribution valuation problem in Section~\ref{sec:setting}; that is, $P(s) := p(s \,|\, \{ \mathcal{D}_s \})$.

The generalized Bayesian loss $L$ in Eq.~\eqref{eq:gbv} is broadly defined, and there are various choices for such a function. For instance, $L$ can be implemented by a sophisticated neural network trained using both $\mathcal{D}_s$ and $\mathcal{D}^*$. However, this would be computationally expensive and training such a neural network would also be ineffective if $\mathcal{D}_s$ is small, as is often the case in applications of data distribution valuation. Thus, for our GBV method, we shall use a more data and computationally efficient approach to construct $L$: the transferability measures in transfer learning~\citep{nguyen2020leep, you2021logme, gholami2023etran}.

\subsection{Negative Transferability as Generalized Bayesian Loss}
\label{sec:bayes-loss}

A transferability measure is a computationally efficient function that quantifies the effectiveness of transfer learning~\citep{you2021logme, gholami2023etran, nguyen2023simple}. Formally, consider a pre-trained source model $m$ and a target task specified by a target training set $D$ drawn from a data distribution $\mathbb{P}(X, Y)$. Let $m_{\mathrm t}$ be the model after running transfer learning from $m$ to $D$. Following \cite{nguyen2023simple}, a transferability measure for classification can be defined as follows.

\begin{definition}
\label{def:trans}
A (perfect) transferability measure is a function $T(m, D) \in \mathbb{R}$ such that ${ T(m, D) \le T(m', D') }$ if and only if ${ \mathbb{P} (y = m_{\mathrm t}(x)) \le \mathbb{P}' (y = m'_{\mathrm t} (x)) }$, where $m$ and $m'$ are source models, $D \sim \mathbb{P}$ and $D' \sim \mathbb{P}'$ are target tasks, $m_{\mathrm t}$ is the model transferred from $m$ to $D$, and $m'_{\mathrm t}$ is the model transferred from $m'$ to $D'$.
\end{definition}

From this definition, a transferability measure can be used as a proxy for test accuracy to compare different source models or target tasks. In practice, the above ideal condition rarely holds, and existing transferability measures only try to approximate it by improving the correlations between $T(m, D)$ and $\mathbb{P} (y = m_{\mathrm t} (x))$.

There are various ways to construct a transferability measure. LEEP~\citep{nguyen2020leep} estimates the transferability using the average loglikelihood $\log p(y | m(x))$ of a simple classifier that makes prediction based on the empirical distribution between the pseudo source label $m(x)$ and the target label $y$. LogME~\citep{you2021logme} relaxes the requirement of a classification head in the model $m$ and leverages a Bayesian approach to estimate the transferability through $\log p(y|m_{\mathrm{f}}(x))$, where $m_{\mathrm{f}}(x)$ is the feature vector extracted by the source model $m$. Energy-based methods, such as ETran~\citep{gholami2023etran}, can also be used to estimate transferability.

For our GBV approach, we propose to use transferability measures to compute the generalized Bayesian loss $L(s, \mathcal{D}_s, \mathcal{D}^*)$. Specifically, we first choose a transferability measure $T$ and construct a source model $m = \phi(s, \mathcal{D}_s, \mathcal{D}^*)$ as well as a target training set $D = \varphi(s, \mathcal{D}_s, \mathcal{D}^*)$ that can be used as arguments for $T$. Here, $\phi$ is a computationally inexpensive procedure to obtain a source model $m$ from $\mathcal{D}_s$ and $\mathcal{D}^*$. For instance, $\phi$ could simply fine-tune a pre-trained model on $\mathcal{D}_s$ to obtain $m$. Similarly, $\varphi$ is an efficient procedure to construct a new target dataset $D$ from $\mathcal{D}_s$ and $\mathcal{D}^*$. In practice, the construction of $\phi$ and $\varphi$ often depends on each specific application. We will detail the choices of $\phi$ and $\varphi$ for some applications in Section~\ref{sec:apps}.

After constructing $\phi$ and $\varphi$, we can use them together with the transferability measure $T$ to build the generalized Bayesian loss:
\begin{align}
L(s, \mathcal{D}_s, \mathcal{D}^*) &= - \frac{1}{\tau} \, T \big( \phi(s, \mathcal{D}_s, \mathcal{D}^*), \varphi(s, \mathcal{D}_s, \mathcal{D}^*) \big),
\label{eq:f-func}
\end{align}
where $\tau > 0$ is a hyper-parameter that balances the effects of $T$ and the prior. We note that Eq.~\eqref{eq:f-func} is a reasonable loss function since smaller loss indicates higher transferability, which also means better compatibility between $\mathcal{D}_s$ and $\mathcal{D}^*$. With the loss~\eqref{eq:f-func}, the posterior in Eq.~\eqref{eq:gbv} can be rewritten as:
\begin{equation}
    p(s \,|\, \{ \mathcal{D}_s \}) \,\propto\, p(s) \exp \hspace{-0.5mm} \big[ \frac{T ( \phi(s, \mathcal{D}_s, \mathcal{D}^*), \varphi(s, \mathcal{D}_s, \mathcal{D}^*) )}{\tau} \big].
    \label{eq:gbv1}
\end{equation}

Our GBV method will return this final posterior as the solution $P(s)$ for the data distribution valuation problem. In practice, if $\mathcal{S}$ is finite, we can compute and store $p(s \,|\, \{ \mathcal{D}_s \})$ directly. If $\mathcal{S}$ is uncountable, we can approximate $p(s \,|\, \{ \mathcal{D}_s \})$ using variational inference~\citep{blei2017variational} or sample from $p(s \,|\, \{ \mathcal{D}_s \})$ using Monte Carlo methods~\citep{rubinstein2016simulation} to compute the training loss~\eqref{eq:weighted_loss}.

In principle, we can tune $\tau$ using the accuracy on the validation set. However, in this work, we propose to use the following ``quick'' value: $\tau = 1/\log_2(\mathcal{\vert S \vert})$ if $\mathcal{S}$ is finite. This is inspired by~\citet{xiao2025sample} when they needed to set a similar scaling factor for evaluating the difficulty of data samples. Our experiment results in Section~\ref{sec:exp} confirm that this is a good choice for $\tau$.

\minisection{Theoretical Property.} From Eq.~\eqref{eq:gbv1} and Def.~\ref{def:trans}, we can easily derive conditions on the posterior and prior to compare the expected accuracies of two target models, assuming a perfect transferability measure. These conditions are stated below (proof in appendix).

\begin{property}
\label{prop:cond}
Consider ${ \{ s_1, s_2 \} \subseteq \mathcal{S} }$ and for ${ i \in \{ 1, 2 \} }$, let ${m_i = \phi(s_i, \mathcal{D}_{s_i}, \mathcal{D}^*)}, \allowbreak {D_i = \varphi(s_i, \mathcal{D}_{s_i}, \mathcal{D}^*)}$, $m_i^{\mathrm t}$ be the target model transferred from $m_i$ to $D_i$, and $\mathbb{P}_i$ be the true data distribution generating $D_i$. Let $p(s)$ and $P(s)$ be the prior and solution of GBV respectively with a perfect transferability measure $T$. We have: \\[5pt]
(a) ${ \mathbb{P}_1 (y = m_1^{\mathrm t}(x)) \leq \mathbb{P}_2 (y = m_2^{\mathrm t}(x)) }$ if and only if $P(s_1)/p(s_1) \le P(s_2)/p(s_2)$. \\[5pt]
(b) As a result, if $p(s)$ is uniform, ${ D_1 = D_2 = \mathcal{D}^* }$, and $P(s_1) \leq P(s_2)$, then $\mathbb{P}^* (y = m_1^{\mathrm t}(x)) \leq \mathbb{P}^* (y = m_2^{\mathrm t}(x))$.
\end{property}

We note that Property~\ref{prop:cond}(b) specifies a sufficient condition for the order-preserving property of GBV that is desirable in previous data distribution valuation work~\citep{xu2024data}. As we will show in Section~\ref{sec:apps}, the condition $D_1 = D_2 = \mathcal{D}^*$ is satisfied for our GBV solution of the annotator evaluation problem.

\begin{table*}[t]
\centering
\caption{Summary of GBV components for two problems in Section~\ref{sec:apps} with a general transferability measure $T$.}
\label{tab:components}
\vspace{2mm}
\resizebox{\textwidth}{!}{%
\begin{tabular}{cll}
\toprule
Component & Annotator evaluation & Data augmentation \\
\midrule
$\mathcal{S}$ & $\{ s_1, \ldots, s_M \}$ (set of annotators) & $\{ s_i \}$, where $s_i = (\psi_i, \alpha_i)$ (set of augmentors) \\
$\mathcal{D}_s$ & $\mathcal{D}_s$ (given noisy sample set from $s$) & $\{ (s(x), y) \,|\, (x, y) \in \mathcal{D}^{\mathrm{tr}} \}$ (transformed training set using $s$) \\
$\mathcal{D}^*$ & $\mathcal{D}^*$ (given validation set) & $\mathcal{D}^{\mathrm{tr}}$ (given training set) \\
$\phi(s, \mathcal{D}_s, \mathcal{D}^*)$ & $m_s$ (small model trained on $\mathcal{D}_s$) & $m^u$ (pre-trained universal model) \\
$\varphi(s, \mathcal{D}_s, \mathcal{D}^*)$ & $\mathcal{D}^*$ (validation set) & $\mathcal{D}_s$ (transformed training set using $s$) \\
$L(s, \mathcal{D}_s, \mathcal{D}^*)$ & $- T(m_s, \mathcal{D}^*) / \tau$ & $- T(m^u, \mathcal{D}_s) / \tau$ \\
\bottomrule
\end{tabular}
}
\end{table*}

\subsection{Specific Instances of GBV}
\label{sec:apps}

So far we have only described GBV generally without specific details on how to construct $\phi$ and $\varphi$. In practice, these functions often depend on each application. In this section, we propose some choices for these functions for the annotator evaluation and the data augmentation problems. The first problem was considered in~\citet{xu2024data}, while the second problem is a novel application of our GBV approach.

\minisection{$\bullet$ Annotator Evaluation.}
In this problem, ${ \mathcal{S} = \{ s_1, s_2, \ldots, s_M \} }$, where each $s_i$ is an annotator who labels training data. The quality of different annotators may vary, with $s_i$ having a noise probability $\epsilon_i = \mathbb{P}(\tilde{y} \neq y \,|\, x, y)$, where $y$ and $\tilde{y}$ are respectively the true and annotated labels of the input $x$. The model trainer has access to a validation set $\mathcal{D}^*$ drawn from the true test data distribution $\mathbb{P}^*$, together with sample sets $\mathcal{D}_{s_i}$ provided by the annotators. The trainer needs to evaluate the annotators using $\mathcal{D}^*$ and $\mathcal{D}_{s_i}$.

To apply GBV to this problem, for each $s \in \mathcal{S}$, we propose $\phi(s, \mathcal{D}_s, \mathcal{D}^*)$ to train a small model $m_s$ using $\mathcal{D}_s$, while $\varphi(s, \mathcal{D}_s, \mathcal{D}^*)$ can simply return $\mathcal{D}^*$. Thus, the generalized Bayesian loss~\eqref{eq:f-func} becomes ${ L(s, \mathcal{D}_s, \mathcal{D}^*) = - T(m_s, \mathcal{D}^*) / \tau }$. Intuitively, the term $T(m_s, \mathcal{D}^*)$ measures the transferability between the noisy model $m_s$ and the validation set $\mathcal{D}^*$, which indicates the quality of annotator $s$ with respect to $\mathbb{P}^*$.

\minisection{$\bullet$ Data Augmentation.}
As another novel contribution, we show that GBV can also be applied to data augmentation~\citep{cubuk2019autoaugment, cubuk2020randaugment, muller2021trivialaugment}. This problem aims to enhance model training by augmenting a training set $\mathcal{D}^{\mathrm{tr}}$ with a set of data augmentors, where each augmentor $(\psi, \alpha)$ is composed of a label-preserving transformation operator $\psi$ and a real-valued magnitude $\alpha \in \mathbb{R}$. For instance, $\psi$ could be the image Gaussian blur operator, and $\alpha$ is the standard deviation of noise. Simple techniques that uniformly sample an augmentor to apply to each training data point perform well in practice \citep{cubuk2020randaugment, muller2021trivialaugment}, but they are not optimal since not all augmentors are beneficial and some may even adversely affect model performance. Thus, instead of uniform sampling, we can improve training by finding a better distribution over the augmentors.

Interestingly, this application can be posed as a data distribution valuation problem where $\mathcal{S} = \{ (\psi_i, \alpha_i) \}$ is the set of all augmentors. Here $\mathcal{S}$ is uncountable since $\alpha_i \in \mathbb{R}$. Given the training set $\mathcal{D}^{\mathrm{tr}}$, we need to find a distribution $P(s)$ over $s = (\psi, \alpha) \in \mathcal{S}$ and use this distribution in the following loss to train the model:
\begin{equation}
\ell_{\text{aug}}(\theta) = \mathbb{E}_{s \sim P(s)} \Big[ \textstyle{\sum_{(x, y) \in \mathcal{D}^{\mathrm{tr}}}} ~ \ell (\theta; s(x), y) \Big],
\label{eq:aug_loss}
\end{equation}
where $s(x)$ is the new input obtained by applying the augmentor $s$ to $x$. Note that this training loss is practically equivalent to~\eqref{eq:weighted_loss} since we often sample $s \sim P(s)$ and compute $s(x)$ on demand during training, and thus do not need the factor $1/n_s$ here.

To find $P(s)$ using our GBV approach, we use $\mathcal{D}^{\mathrm{tr}}$ as the validation set, i.e., $\mathcal{D}^* = \mathcal{D}^{\mathrm{tr}}$, and construct the sample set $\mathcal{D}_s$ by applying the augmentor $s$ to each data point in $\mathcal{D}^{\mathrm{tr}}$, i.e., $\mathcal{D}_s = \{ (s(x), y) \,|\, (x, y) \in \mathcal{D}^{\mathrm{tr}} \}$. Now if we train a model on $\mathcal{D}_s$ and compute its transferability to $\mathcal{D}^*$ as in the previous application, this value may not be useful since it just captures the magnitude $\alpha$ of the augmentor $s$. Thus, for this problem, we propose to simply let the function $\phi(s, \mathcal{D}_s, \mathcal{D}^*)$ ignore both $\mathcal{D}_s$ and $\mathcal{D}^*$ and just return a universal model $m^u$, such as a pre-trained model on ImageNet. Additionally, we let the function $\varphi(s, \mathcal{D}_s, \mathcal{D}^*)$ return the sample set $\mathcal{D}_s$. In this case, the generalized Bayesian loss~\eqref{eq:f-func} becomes $L(s, \mathcal{D}_s, \mathcal{D}^*) = - T(m^u, \mathcal{D}_s) / \tau$. Intuitively, the term $T(m^u, \mathcal{D}_s)$ measures the similarity between a well-trained universal model and $\mathcal{D}_s$, which can tell us how the augmentor $s$ affects the original training set $\mathcal{D}^{\mathrm{tr}}$ universally.

\minisection{Remarks.} In Table~\ref{tab:components}, we summarize all the above GBV components. We note that from Property~\ref{prop:cond}(b), with the uniform prior and a perfect transferability measure, GBV is theoretically order-preserving in the above annotator evaluation application. Besides, we keep the choice of the transferability measure $T$ flexible and will show in the experiments that different measures can all lead to improved model accuracy.

\subsection{Relaxation on the Validation Set}
\label{sec:relaxation}

Although our GBV method above assumes the existence of a labeled validation set $\mathcal{D}^*$, this assumption can be relaxed so that GBV can operate even when $\mathcal{D}^*$ is unlabeled or when $\mathcal{D}^*$ is not available. The former case can be naturally handled by leveraging a label-free transferability measure, such as the energy part of ETran~\citep{gholami2023etran}. The latter case can be addressed by using $\mathcal{D}^* = \bigcup_{s \in \mathcal{S}} \mathcal{D}_s$, i.e., we treat the union of all sample sets as our reference set. This setting of $\mathcal{D}^*$ corresponds to the target distribution $\mathbb{P}^* = \sum_{s \in \mathcal{S}} \mathbb{P}_s$, the uniform mixture over all data sources' distributions, which is worst-case optimal in a game-theoretic perspective~\citep{xu2024data}.

\section{CONTINUAL DATA DISTRIBUTION VALUATION}
\label{sec:cbdsw}

One major advantage of using generalized Bayesian inference for GBV is that it can naturally handle the continual setting, as analogous to Bayesian continual learning~\citep{nguyen2018variational, nguyen2024lifelong}. In the continual setting, the model trainer still has the validation set $\mathcal{D}^*$ but will receive a subset $\mathcal{D}_{s,t}$ of the source sample set $\mathcal{D}_s$ sequentially over several time steps ${ t = 1, 2, \ldots, T }$, such that $\mathcal{D}_s = \cup_t \mathcal{D}_{s,t}$. Similar to continual learning, at each time $t$, the model trainer needs to update the solution $P(s)$ using only the subsets $\{ \mathcal{D}_{s,t} : s \in \mathcal{S} \}$, without access to the subsets from previous time steps.

This continual setting is useful in scenarios where the model trainer is not allowed to keep past data, e.g., due to policy or privacy constraints. For instance, in annotator evaluation, the annotators may require the trainer to destroy their past sample data after a certain time period. Besides, the label quality of each annotator may change over time, so the model trainer may request a new batch of sample data from the annotators for re-evaluation. This scenario can be considered a continual data distribution valuation problem where the model trainer needs to continuously update $P(s)$ using the new sample data.

We can extend GBV to this setting using the Bayesian principle. For any time $t$ and data source $s$, let ${ \mathcal{D}_{s,{1:t}} = \cup_{i=1}^t \mathcal{D}_{s,i} }$. Using the short-handed notations as in Section~\ref{sec:gbv}, we can derive a recursive expression for the posterior $p(s \,|\, \{ \mathcal{D}_{s,{1:t}} \})$ as follows:
\begin{align*}
    p(s \,|\, \{ \mathcal{D}_{s,{1:t}} \}) &\propto p(s) \, p(\{ \mathcal{D}_{s,{1:t}} \} \,|\, s)  \\
    &= p(s) \, p(\{ \mathcal{D}_{s,{1:t-1}} \} \,|\, s) \, p(\{ \mathcal{D}_{s,t} \} \,|\, s) \\
    &\propto p(s \,|\, \{ \mathcal{D}_{s,{1:t-1}} \}) \, p(\{ \mathcal{D}_{s,t} \} \,|\, s).
\end{align*}
Thus, if we let $P_{t-1}(s) = p(s \,|\, \{ \mathcal{D}_{s,{1:t-1}} \})$ be the GBV solution at time step $t-1$ and use the generalized Bayesian loss $L(s, \mathcal{D}_{s,t}, \mathcal{D}^*)$ instead of $p(\{ \mathcal{D}_{s,t} \} \,|\, s)$, we can rewrite the expression above as:
\begin{align}
    P_t(s) &\propto P_{t-1}(s) \exp [ - L(s, \mathcal{D}_{s,t}, \mathcal{D}^*) ] \notag \\
    &\hspace{-0.8cm} = P_{t-1}(s) \exp \hspace{-0.5mm} \big[ \frac{T ( \phi(s, \mathcal{D}_{s,t}, \mathcal{D}^*), \varphi(s, \mathcal{D}_{s,t}, \mathcal{D}^*) )}{\tau} \big].
    \label{eq:cgbv}
\end{align}
We call methods that use Eq.~\eqref{eq:cgbv} to solve the continual data distribution valuation problem the \emph{Continual Generalized Bayes Valuation} (CGBV) methods. An advantage of CGBV is that it can reduce forgetting the previous information in $\mathcal{D}_{s,1:t-1}$, even without access to those data in the current step $t$. This is similar to the effects of Bayesian methods on catastrophic forgetting in continual learning \citep{nguyen2018variational, nguyen2024lifelong}.

\section{EXPERIMENTS}
\label{sec:exp}

In this section, we empirically evaluate GBV and CGBV on the annotator evaluation and data augmentation problems. All experiments were conducted on a system with four NVIDIA V100 GPUs. More details of our experiment settings are given in the appendix.

\subsection{Annotator Evaluation}
\label{sec:exp_annotator}

We first experiment with the annotator evaluation problem on two widely used image classification datasets: CIFAR-10~\citep{krizhevsky2009learning} and CUB-200-2011~\citep{wah2011caltech}. We use the original train-test splits for both datasets and set up the experiment as follows.

$\bullet$ {\bf CIFAR-10.} This dataset contains 50,000 training and 10,000 test images across 10 classes. We use 100 random test images per class as the validation set $\mathcal{D}^*$ and distribute the training images among 5 annotators, each of whom will label 10,000 training images. Following \citet{xu2024data}, we set the label noise probability $\epsilon_i = i/5$ for each annotator $i \in \{0,1,2,3,4\}$ and randomly corrupt each label based on this probability. For the sample set $\mathcal{D}_s$, we use 100 random images per class from the respective noisy training set of each annotator.

$\bullet$ {\bf CUB-200-2011.} This is a more challenging dataset that comprises 11,788 labeled bird images from 200 species (5,994 for training and 5,794 for testing). There are around 30 images per class for training and roughly the same number of images per class for testing. We randomly select 10 test images per class to construct $\mathcal{D}^*$ and distribute the training set among 3 annotators, each of whom will label around 10 images per class. We also randomly corrupt the labels with noise probability $\epsilon_i = i/3$ for each annotator $i \in \{0,1,2\}$. Since the training set from each annotator is small, we use this whole set as $\mathcal{D}_s$.

\begin{table}[t!]
\centering
\caption{Test accuracy (\%) of different methods for annotator evaluation. Bold numbers and asterisks (*) indicate the best and second best accuracies on each dataset respectively. Our GBV method outperforms the baselines on both datasets.}
\label{tab:gbv_anno}
\vspace{3mm}
\small
\begin{tabular}{lcc}
\toprule
\multirow{2}{*}{Method} & \multicolumn{2}{c}{Dataset}\\
\cmidrule{2-3}
 & CIFAR-10 & CUB-200-2011 \\
\midrule
Uniform & 74.37 $\pm$ 0.23 & 52.45 $\pm$ 0.85 \\ 
DAVINZ~\citeyearpar{wu2022davinz} & 74.82 $\pm$ 0.62 & 52.06 $\pm$ 0.38 \\ 
LAVA~\citeyearpar{just2023lava} & 74.74 $\pm$ 0.91 & - \\
MMD~\citeyearpar{xu2024data} & 75.95 $\pm$ 0.91 & 52.62 $\pm$ 0.80 \\
\midrule
GBV (quick $\tau$) & 77.94 $\pm$ 0.08* & 57.58 $\pm$ 0.96* \\
GBV (best $\tau$) & \textbf{78.32 $\pm$ 0.38} & \textbf{58.08 $\pm$ 0.32} \\
\bottomrule
\end{tabular}
\end{table}

We run GBV with the uniform prior and settings in Section~\ref{sec:apps} for the annotator evaluation problem. For CIFAR-10, we use the ResNet-18 backbone~\citep{he2016deep} and train the models $m_s$ with stochastic gradient descent. We choose LEEP as the transferability measure due to its known stability~\citep{kazemi2025benchmarking}, and employ the quick $\tau$ discussed in Section~\ref{sec:bayes-loss} (i.e., $\tau = 1/\log_2(5) \approx 0.43$). After obtaining the solution $P(s)$ from GBV, we train the final model with the loss \eqref{eq:weighted_loss} using Adam~\citep{kingma2014adam}. For CUB-200-2011, we follow the same settings but use the ResNet-34 backbone due to its good accuracy on this dataset. In all experiments, we initialize our models with pre-trained weights on ImageNet.

We compare our GBV method with 4 baselines: Uniform, DAVINZ~\citep{wu2022davinz}, LAVA~\citep{just2023lava}, and MMD~\citep{xu2024data}. The Uniform baseline simply sets $P(s)$ to the uniform distribution when training the final model. DAVINZ and LAVA are two strong baselines for traditional data valuation with a reference set, while MMD is the most recent data distribution valuation method without a central broker. 

We evaluate the performance of all baselines based on the official implementations provided in their source repositories. Since LAVA is an instance-wise valuation framework, we derive an aggregated distribution value for each annotator by averaging the scores of all data points within their sample set. As the labeled reference set is available in our setting, we adopt conditional-MMD, the state-of-the-art approach for data distribution valuation, as a strong baseline for comparison. For all baselines, the resulting scores are passed through a softmax function to produce a valid distribution that is consistent with the training procedure used in our method. This unified protocol ensures a fair and comparable evaluation across all methods. Full details on the baseline configurations are provided in the appendix.

From the results in Table~\ref{tab:gbv_anno}, using uniform weights or traditional data valuation methods (DAVINZ and LAVA) results in lower test accuracies on both datasets. This is likely because data valuation methods only evaluate individual data points, making them unsuitable for distribution valuation. Furthermore, LAVA fails to generate valuation scores on CUB-200-2011 within a reasonable time frame due to the high computational cost of optimal transport on many classes.\footnote{The computational complexity of class-wise Wasserstein distance used in LAVA is $\mathcal{O}(Cn^2)$ per iteration, where $n$ is the number of samples and $C$ is the number of classes. This complexity makes LAVA unsuitable for fine-grained classification tasks that involve a large number of classes.} In contrast, distribution valuation approaches (MMD and GBV) both outperform the above baselines, with our GBV method having better accuracy than MMD. Compared to the best $\tau$ (obtained by grid search), GBV with the quick $\tau$ value is still very competitive and achieves the second best accuracy on both datasets. 

In the appendix, we provide additional results for evaluating the correlations between our proposed method's valuations and actual test accuracies on both datasets. The results show that GBV consistently outperforms the strongest baseline, MMD, across both standard and relaxed validation set settings in Section~\ref{sec:relaxation}.

\begin{table}[t!]
\centering
\caption{Test accuracy (\%) of different methods for data augmentation. Bold numbers and asterisks (*) indicate the best and second best accuracies on each dataset respectively. Our GBV method outperforms the baselines on both datasets.}
\label{tab:gbv_aug}
\vspace{3mm}
\resizebox{\columnwidth}{!}{%
\begin{tabular}{lcc}
\toprule
\multirow{2}{*}{Method} & \multicolumn{2}{c}{Dataset}\\
\cmidrule{2-3}
& CUB-200-2011 & Stanford-Dogs \\
\midrule
AutoAugment~\citeyearpar{cubuk2019autoaugment} & 67.30 $\pm$ 0.39 & 65.64 $\pm$ 0.58 \\ 
RandAugment~\citeyearpar{cubuk2020randaugment} & 67.73 $\pm$ 0.23 & 67.12 $\pm$ 0.31 \\ 
TrivialAugment~\citeyearpar{muller2021trivialaugment} & 69.43 $\pm$ 0.49 & 67.64 $\pm$ 0.23 \\ 
EntAugment~\citeyearpar{yang2024entaugment} & 72.38 $\pm$ 0.40 & 72.17 $\pm$ 0.27 \\ 
SRA~\citeyearpar{xiao2025sample} & 66.53 $\pm$ 0.42 & 66.26 $\pm$ 0.42 \\
\midrule
GBV (quick $\tau$) & 73.24 $\pm$ 0.39* & 73.16 $\pm$ 0.43* \\
GBV (best $\tau$) & \textbf{73.92 $\pm$ 0.31} & \textbf{73.20 $\pm$ 0.26} \\
\bottomrule
\end{tabular}
}
\end{table}

\subsection{Data Augmentation}
\label{sec:exp_aug}

We next consider data augmentation on two fine-grained visual recognition datasets: CUB-200-2011 \citep{wah2011caltech} and Stanford-Dogs \citep{khosla2011novel}. The former is the same as in the previous experiment, while the latter contains 20,580 images of 120 dog breeds. We use the original train-test splits for both datasets and set up the experiment as follows.

For each dataset, we use its training set $\mathcal{D}^{\mathrm{tr}}$, enhanced with a set of data augmentors $\mathcal{S}$, to train a model. The set $\mathcal{S}$ includes various PyTorch operators~\citep{paszke2019pytorch} such as Rotation, AutoContrast, and Brightness (see appendix for the full list). If an operator $\psi$ has a continuous magnitude, for simplicity, we discretize its range into 5 distinct values $\{\alpha_i\}_{i=1}^{5}$, resulting in 5 different augmentors $s_i = (\psi, \alpha_i)$ that share the same operator $\psi$. We run GBV with the uniform prior and settings in Section~\ref{sec:apps} for data augmentation, where we choose the universal model $m^u$ as a ResNet-34 pre-trained on ImageNet since it is a good model on these datasets. As with the previous experiment, we also use the LEEP transferability measure and the quick value of $\tau$.
After obtaining the solution $P(s)$ from GBV, we use it to sample the augmentors when training the final model with the loss~\eqref{eq:aug_loss}, as usually done in previous data augmentation work \citep{muller2021trivialaugment, yang2024entaugment}. We train this final model by fine-tuning an ImageNet pre-trained ResNet-34. 

\begin{figure}
\centering
\includegraphics[width=0.8\columnwidth]{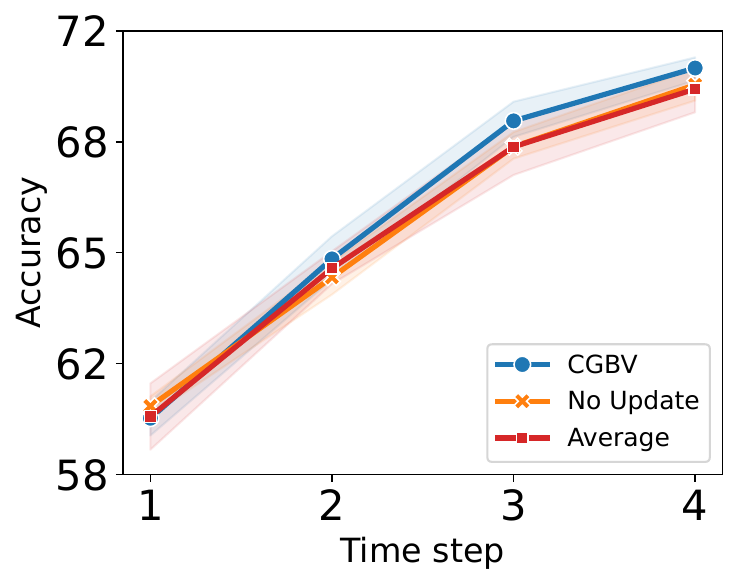}
\vspace{-2mm}
\caption{Test accuracy of different methods for continual annotator evaluation. Our CGBV method consistently outperforms the baselines on this problem.}
\label{fig:continual}
\vspace{-2mm}
\end{figure}

We compare our GBV method with 5 baselines: AutoAugment~\citep{cubuk2019autoaugment}, RandAugment~\citep{cubuk2020randaugment}, TrivialAugment~\citep{muller2021trivialaugment}, EntAugment~\citep{yang2024entaugment}, and SRA~\citep{xiao2025sample}. These are chosen because they are policy-free methods as our approach. AutoAugment offers a fixed pre-learned set of augmentors.
RandAugment and TrivialAugment assume a uniform distribution over the augmentors, making them a direct baseline for our method. EntAugment and SRA employ sample-aware strategies, which dynamically adjust the magnitude of the augmentors.

The results in Table~\ref{tab:gbv_aug} indicate that directly applying AutoAugment yields suboptimal performance because this method was developed for ImageNet and not tailored to specific characteristics of our target domains. While RandAugment and TrivialAugment show improvements, they are both worse than EntAugment. Notably, despite being recent, SRA has low performance on both datasets. Compared to these baselines, our GBV method is consistently better on both datasets, and using the quick $\tau$ value is also highly competitive to using the best $\tau$ value.\footnote{On CUB-200-2011, GBV achieves comparable accuracy to previous work~\citep{zhang2018fine} that fine-tunes a model pre-trained on a subset of bird images from ImageNet.}

\subsection{Continual Annotator Evaluation}
\label{sec:exp_cont_anno}

In this experiment, we evaluate the performance of the CGBV method on the continual annotator evaluation problem. Specifically, we modify the CIFAR-10 experiment in Section~\ref{sec:exp_annotator} and allow the annotators to provide sample sets $\mathcal{D}_{s,t}$ over 4 time steps. At every step, each annotator provides 50 labeled samples per class. We adopt similar noise probabilities from the previous experiment, but randomly permute these probabilities among the annotators at each step. This ensures that the quality of each annotator varies over time, which necessitates continuous updates of $P(s)$. At every step, the available training set from each annotator contains 100 samples per class. This set will be merged with training data from previous steps to form the new training set to train the model.

We run CGBV with the initial uniform prior as $P_0(s)$ and update $P_t(s)$ over time using Eq.~\eqref{eq:cgbv} with the LogME transferability measure. We keep the same CIFAR-10 settings as in Section~\ref{sec:exp_annotator}, except that we train the models $m_s$ for 20 epochs. To benchmark our method, at every step $t = 1, \ldots, 4$, we compare it with two baselines: \textit{No Update}, which uses only the first distribution $P_1(s)$; and \textit{Average}, which uses the average distribution across all steps, i.e., $\sum_{1 \le i \le t} P_i(s) /t$.

The result of this experiment is reported in Figure~\ref{fig:continual}. From the figure, training without updating $P_t(s)$ or with the average distribution leads to consistently lower accuracy compared to CGBV. This result highlights the effectiveness of our method in dynamically estimating the data distribution values in a streaming scenario.

\subsection{Comparison of Running Time}

\begin{table}[t]
\centering
\footnotesize
\caption{Running time of different methods.}
\label{tab:runtime}
\vspace{2mm}
\begin{tabular}{cc|cc}
\toprule
Method & Time (s) & Method & Time (s) \\
\midrule
DAVINZ & 10.68 $\pm$ 0.4 & MMD & 1.05 $\pm$ 0.58 \\ 
LAVA & 370.21 $\pm$ 5.37 & GBV & 0.85 $\pm$ 0.17 \\
\bottomrule
\end{tabular}
\end{table}

We also compare the running time of GBV to the baselines on the CIFAR-10 annotator evaluation experiment in Section~\ref{sec:exp_annotator}. In Table~\ref{tab:runtime}, we report the running time (in seconds) per annotator for each method. From the table, the data distribution valuation methods (GBV and MMD) are significantly more efficient than the data valuation methods (DAVINZ and LAVA). Furthermore, GBV is also slightly faster than MMD. This highlights the efficiency and practicality of our method.

\begin{figure}[t]
    \centering
    \begin{minipage}{0.49\columnwidth}
        \centering
        \includegraphics[width=\linewidth]{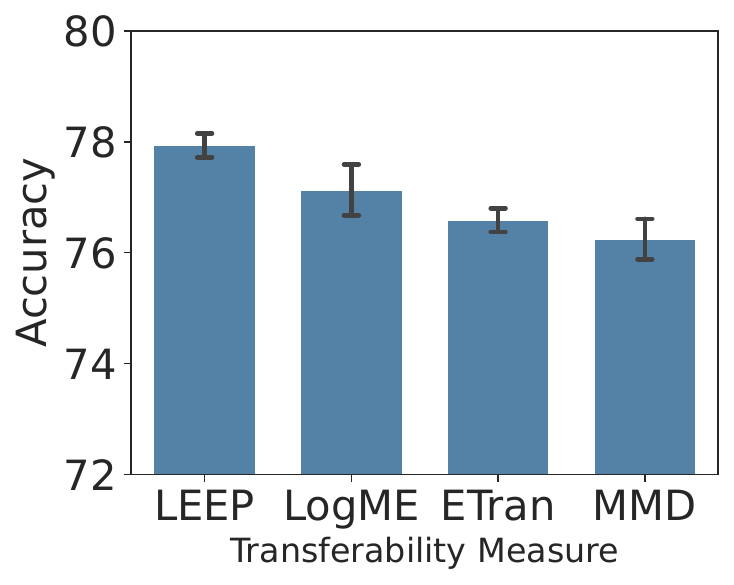}
    \end{minipage}
    \hfill 
    \begin{minipage}{0.46\columnwidth}
        \centering
        \includegraphics[width=\linewidth]{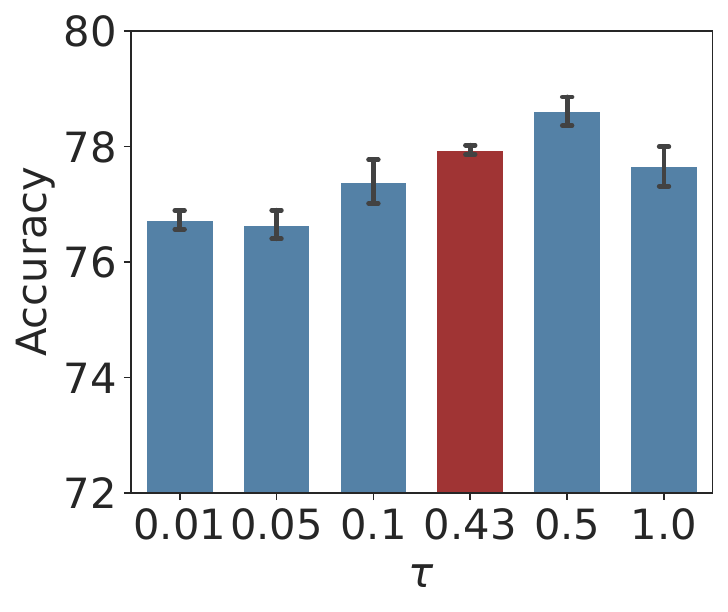}
    \end{minipage}
    \vspace{-1mm}
    \caption{Effects of different transferability measures (left) and $\tau$ (right) on test accuracy. The red column on the right is for the quick $\tau$ value.}
    \label{fig:ablation_studies}
\end{figure}

\subsection{Ablation Studies}
\label{sec:abla}

In this section, we conduct the following ablation studies using the CIFAR-10 experiment in Section~\ref{sec:exp_annotator}.

\minisection{Effects of Transferability Measures.}
We examine the performance of GBV with respect to 3 different transferability measures: LEEP~\citep{nguyen2020leep}, LogME~\citep{you2021logme} and ETran~\citep{gholami2023etran}, which are known to be robust and stable in other applications~\citep{kazemi2025benchmarking}. We add MMD~\citep{xu2024data} to the comparison since it can also be used in GBV as a transferability measure. As shown in Figure~\ref{fig:ablation_studies}, varying the transferability measures results in accuracies from 75.95\% (MMD) to 77.93\% (LEEP), which are all better than the baselines in Table~\ref{tab:gbv_anno}.

\minisection{Effects of $\tau$.} We then evaluate the sensitivity of GBV to $\tau$. When varying $\tau$ from 0.01 to 1.0, Figure~\ref{fig:ablation_studies} (right) shows the robustness of GBV, with accuracies ranging from 76.64\% to 78.48\%, which are all better than the baselines in Table~\ref{tab:gbv_anno}. Nevertheless, tuning this hyper-parameter can be beneficial, e.g., setting $\tau=0.5$ yields a distinct improvement over $\tau=1$. Notably, using the quick $\tau$ (red column) achieves a strong performance without the cost of hyper-parameter search.

\begin{wrapfigure}{r}{0.45\columnwidth}
{\vskip -5mm}
\includegraphics[width=\linewidth]{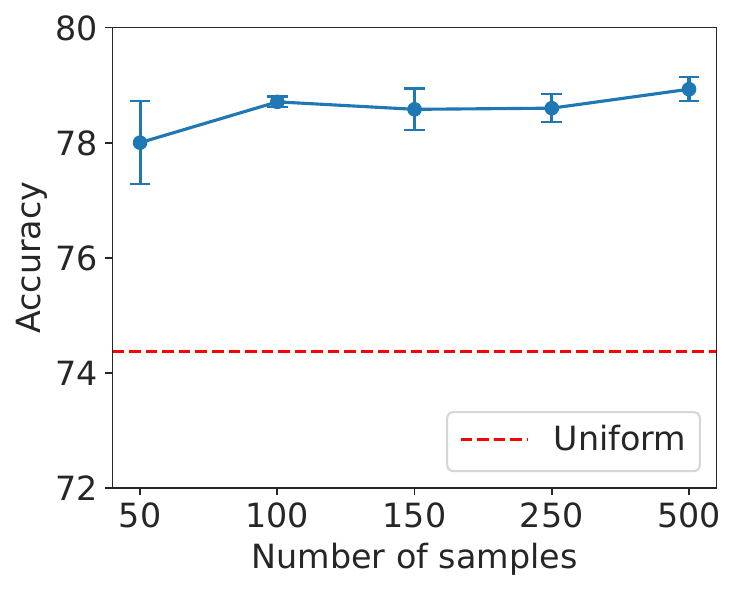}
{\vskip -3mm}
\caption{Test accuracy of GBV with respect to sample size.}
\label{fig:abla_size}
{\vskip -2mm}
\end{wrapfigure}

\minisection{Effects of Sample Size.}
We investigate the robustness of GBV to the size of $\mathcal{D}_s$. Figure~\ref{fig:abla_size} shows the accuracies when varying the number of samples per class in $\mathcal{D}_s$ from 50 to 500. Across all sample sizes, GBV consistently outperforms the uniform baseline (red dashed line). It also shows stable performance, with the sample sizes having only a slight effect on our method. Notably, using only 100 samples per class offers a competitive accuracy (78.71\%), and further increasing the sample size does not result in a substantial accuracy improvement. This suggests that a moderate number of samples is sufficient for a reliable performance of our method.

\section{DISCUSSIONS}

When $\mathcal{S}$ is continuous, we may not have a closed form or conjugate prior for the GBV solutions due to the complex form of the transferability measure. This limitation is quite common for real-world applications of Bayesian methods and could be addressed by, e.g., using an approximate distribution.

As another remark, the theoretical property of GBV (Property \ref{prop:cond}) assumes a perfect transferability measure. While this does not affect the usage of GBV in practice, future work could improve the theory by allowing an imperfect transferability measure with some error rate and quantifying how it would affect the posterior. Besides, PAC-Bayes bounds~\citep{alquier2024user} could also be considered for GBV.

Finally, while our method is inherently generalizable, this study focuses exclusively on computer vision tasks. This leaves an open question for investigating the efficacy of GBV on other types of data. We consider these extensions, along with the scaling of GBV to larger datasets, as promising directions for future work.

\section{CONCLUSION}
\label{sec:conclusion}

We proposed GBV, a novel framework that approaches the data distribution valuation problem from a Bayesian perspective, with the generalized Bayesian loss constructed from a transferability measure. Our framework can be applied to practical applications such as annotator evaluation and data augmentation. Leveraging properties of Bayesian inference, we also extended GBV to the streaming setting where data arrive sequentially. Empirical results confirmed the effectiveness of our methods in various settings.

\acknowledgments{This work made use of the facilities of the N8 Centre of Excellence in Computationally Intensive Research (N8 CIR) provided and funded by the N8 research partnership and EPSRC (Grant No. EP/T022167/1). The Centre is coordinated by the Universities of Durham, Manchester, and York. Part of this work was done when the authors were at the Florida International University.}

\bibliography{aistats}
\bibliographystyle{plainnat}

\clearpage
\appendix
\thispagestyle{empty}

\onecolumn

\aistatstitle{Appendix for ``Data Distribution Valuation Using \\ Generalized Bayesian Inference''}

\section{PROOF OF THEORETICAL PROPERTY~\ref{prop:cond}}

(a) From Definition~\ref{def:trans} in the main paper, we have:
\begin{align*}
      \quad \mathbb{P}_1 (y = m_1^{\mathrm t}(x)) \leq \mathbb{P}_2 (y = m_2^{\mathrm t}(x))
     &\Leftrightarrow T(m_1, D_1) \le T(m_2, D_2) \\
     &\Leftrightarrow \exp \left [ \frac{T(m_1, D_1)}{\tau} \right] \le \exp \left [ \frac{T(m_2, D_2)}{\tau} \right]  \\
     &\Leftrightarrow \frac{p(s_1 \,|\, \{ \mathcal{D}_s \})}{p(s_1)} \le \frac{p(s_2 \,|\, \{ \mathcal{D}_s \})}{p(s_2)} \tag{using Eq.~\ref{eq:gbv1}} \\
     &\Leftrightarrow \frac{P(s_1)}{p(s_1)} \le \frac{P(s_2)}{p(s_2)}.
\end{align*}

(b) This is a direct consequence of part (a) with $p(s_1) = p(s_2)$ due to $p(s)$ being uniform, $\mathbb{P}_1 = \mathbb{P}_2 = \mathbb{P}^*$ due to $D_1 = D_2 = \mathcal{D}^*$, and $P(s_1) \leq P(s_2)$.

\section{MORE DETAILS ON TRANSFERABILITY MEASURES}

\minisection{ETran}~\citep{gholami2023etran} estimates model transferability by leveraging energy-based models to measure how well a pre-trained model's extracted features align with the target data distribution. Rather than relying on task-specific output logits, ETran computes free energy directly over feature representations, making the metric task-independent. Specifically, the feature-based free energy is computed as the negative log-sum-exp of the feature components, where lower energy signifies that the sample is in-distribution for the source model. The overall transferability score is then the average negative free energy across the target dataset.

\minisection{LEEP}~\citep{nguyen2020leep} quantifies the transferability between a pre-trained source model and a target dataset by evaluating the alignment between source and target label distributions. This method computes the empirical conditional distribution of the target labels given the source labels and uses this distribution to construct the expected empirical predictor that maps source label predictions to target labels. The LEEP transferability score is defined as the average log-likelihood of this predictor.

\minisection{LogME}~\citep{you2021logme} assesses model transferability by estimating the compatibility between the extracted features of a pre-trained model and the target labels using a gradient-free Bayesian framework. Instead of relying on a single optimized weight, it calculates the logarithm of marginalized likelihood (evidence) by integrating over all possible weights of a linear model under Gaussian assumptions. Due to the conjugate properties of Gaussian distributions, this integral has a closed-form expression that can be optimized using an efficient fixed-point algorithm. The LogME score is the average maximum log-evidence across the target dataset.

\section{MORE DETAILS ON DATA VALUATION BASELINES}

\minisection{DAVINZ}~\citep{wu2022davinz} provides a computationally efficient, training-free alternative to evaluate the value of data used to train a deep network at initialisation. By leveraging the neural tangent kernel (NTK), it derives a domain-aware generalization bound that accounts for distribution shifts between the source and target domains. The score function combines two components: (1) an in-domain complexity term based on the initial prediction error and the NTK matrix; and (2) an out-of-domain discrepancy term that penalizes the domain shift using the kernel mean discrepancy.

\minisection{LAVA}~\citep{just2023lava} is a training-free framework that evaluates the utility of training data through optimal transport (OT). It constructs discrete probability measures from the training and validation sets, which are utilized to estimate the dataset discrepancy via class-wise Wasserstein distance. The value of each individual data point is determined by the sensitivity of the Wasserstein distance to perturbations on probability mass related to that data point. Finally, LAVA employs the entropy-regularized OT via the Sinkhorn algorithm to ensure its scalability to larger datasets.

\minisection{MMD}~\citep{wu2022davinz} estimates data distribution valuations by modeling data source distributions via the Huber heterogeneity model. In this model, each source distribution is a mixture of the true target distribution and a distribution that captures the heterogeneity of this source. The value of a source distribution is defined as its negated maximum mean discrepancy to the ideal target distribution, which can be empirically estimated from data samples.

\section{MORE DETAILS ON EXPERIMENT SETTINGS}

\subsection{Annotator Evaluation in Section~\ref{sec:exp_annotator}}

For CIFAR-10, we use the ResNet-18 backbone~\citep{he2016deep} and train the models $m_s$ for 10 epochs with stochastic gradient descent. Besides using the quick $\tau$ value, we also find the best $\tau \in \{ 0.01, 0.05, 0.1, 0.5, 1.0 \}$ by grid search. After obtaining the solution $P(s)$, we train the final model with the loss \eqref{eq:weighted_loss} using Adam~\citep{kingma2014adam} for 40 epochs. The learning rate is set at $10^{-4}$, and it is linearly decayed by a factor of 10 every 10 epochs after the $20^{th}$ epoch. For CUB-200-2011, we follow the same settings but use the ResNet-34 backbone and the quick  $\tau = 1/\log_2(3) \approx 0.63$. In all experiments, we initialize our models with pre-trained weights on ImageNet. We run all experiments with 5 different random seeds and report the average accuracies together with the standard errors.

\subsection{Data Augmentation in Section~\ref{sec:exp_aug}}

\minisection{Augmentation Space.}
We describe in Table~\ref{tab:aug_ops} the full augmentation space $\mathcal{S}$ used in our data augmentation experiment. A dashed line in the table indicates transformations without a magnitude parameter $\alpha$, for which we fix $\alpha = 0$. 
This augmentation space $\mathcal{S}$ is used consistently across all methods, except for AutoAugment \citep{cubuk2019autoaugment}, whose policies are discovered via a reinforcement learning-based search algorithm that is computationally infeasible to run on our system. Thus, for AutoAugment, we instead use the ImageNet-trained policies for both CUB-200-2011 and Stanford-Dogs to leverage their general applicability.

\begin{table}[t]
\centering
\caption{Data augmentation space used in our experiment.}
\vspace{2mm}
\begin{tabular}{cc|cc}
\toprule
augmentor & range & augmentor & range \\
\midrule
equalise & - & rotate & $-90^\circ$ - $90^\circ$ \\
solarise & 0 - 256 & color & 0.0 - 5.0 \\
posterise & 2 - 8 & contrast & 0.0 - 0.9 \\
brightness & 0 - 5 & sharpness & 0.0 - 5. \\
shear\_x & 0.0 - 0.3 & shear\_y & 0.0 - 0.3\\
translate\_x & 0 - 32 & translate\_y & 0 - 32 \\
auto\_contrast & - & gaussian\_blur & 0.1 - 5.0\\
invert & - & gaussian\_noise & 0.1 - 3.0\\
\bottomrule
\end{tabular}
\label{tab:aug_ops}
\end{table}

\minisection{More Training Details.}
Similar to the previous experiment, we use the quick $\tau = 1/\log_2(68)$ since $|\mathcal{S}| = 68$ for the discretized augmentation space $\mathcal{S}$. The best $\tau$ value is also found by grid search from $\{0.01, 0.05, 0.1, 0.5, 1.0\}$ (best $\tau = 0.05$). The final model is trained by fine-tuning an ImageNet pre-trained ResNet-34 for 100 epochs using the Adam optimizer~\citep{kingma2014adam} to minimize the loss~\eqref{eq:aug_loss}. The initial learning rate is set to $10^{-4}$ and is linearly decayed by a factor of 10 every 10 epochs after the $20^{th}$ epoch. We run all experiments with 5 different random seeds and report the average accuracies together with the standard errors.

\section{MORE EXPERIMENT RESULTS}

\subsection{Correlation with Actual Test Accuracy}
\label{sup:exp_ref}

\begin{wraptable}{r}{0.45\textwidth}
    \centering
    \vspace{-7mm}
    \caption{Pearson correlation coefficients between valuation methods' solutions and test accuracy with a labeled reference set. The best results for each dataset are highlighted in bold.}
    \label{tab:corr_with_ref}
    \vspace{2mm} 
    \begin{tabular}{lcc}
        \toprule
        Method & CIFAR10       & CUB-200-2011  \\
        \midrule
        MMD    & \textbf{0.99} & 0.93          \\
        GBV    & \textbf{0.99} & \textbf{0.98} \\
        \bottomrule
    \end{tabular}
\end{wraptable}

We provide an evaluation of the correlation between GBV valuations and actual test accuracies when the reference set $\mathcal{D}^*$ is fully available. Using the experiment setup in Section~\ref{sec:exp_annotator}, we obtain the GBV solutions with LogME transferability~\citep{you2021logme} and the optimal $\tau$, then compute its Pearson correlation to the accuracies of the models $m_s$ on the test set. As the baseline, we use conditional-MMD~\citep{xu2024data}, the state-of-the-art method for data distribution valuation, with its scores passed through a softmax function to produce a valid distribution. As shown in Table~\ref{tab:corr_with_ref}, GBV correlates better with the test accuracy than MMD. This indicates that GBV produces a more reliable weighting of data sources, which explains the superior performance when training the buyer model.

\subsection{Correlation with Test Accuracy Under Relaxed Validation Set}

\begin{wraptable}{r}{0.45\textwidth}
    \centering
    \vspace{-7mm}
    \caption{Pearson correlation coefficients between valuation methods' solutions and test accuracy on CIFAR-10 with unlabeled and entirely unavailable reference sets. The best results in each setting are highlighted in bold.}
    \label{tab:corr_relaxed}
    \vspace{2mm}
    \begin{tabular}{lcc}
        \toprule
        Method & Unlabeled     & No reference set \\
        \midrule
        MMD    & 0.76          & 0.85             \\
        GBV    & \textbf{0.83} & \textbf{0.87}    \\
        \bottomrule
    \end{tabular}
\end{wraptable}

Using the CIFAR-10 experiment setup in Section~\ref{sec:exp_annotator}, we further evaluate the robustness of GBV under the relaxed validation set settings in Section~\ref{sec:relaxation}, where the reference set is either unlabeled or entirely unavailable. In the first case, we remove all labels from the reference set $\mathcal{D}^*$ and employ a label-free transferability measure, specifically the energy-based component of ETran~\citep{gholami2023etran}, for GBV valuation. In the second case where no reference set is given, we follow \citet{xu2024data} and use $\mathcal{D}^* = \bigcup_{s \in \mathcal{S}} \mathcal{D}_s$ as the aggregated reference set for GBV with LogME transferability~\citep{you2021logme}. In both cases, we compute the posterior using the best $\mathcal{\tau}$ value. To ensure a fair comparison, the MMD baseline is evaluated under the same conditions. We follow the procedure in Section~\ref{sup:exp_ref} and report Pearson correlation coefficients between these methods' valuations and the test accuracy in Table~\ref{tab:corr_relaxed}. The results indicate that GBV consistently surpasses MMD in both settings, underscoring its robustness even in extreme evaluation scenarios.

\subsection{Ablation Study on the Effect of Universal Model for Data Augmentation}

\begin{wraptable}{r}{0.45\textwidth}
    \centering
    \vspace{-7mm}
    \caption{Final test accuracy (\%) for data augmentation on CUB-200-2011 when using different universal models $m^u$ for GBV.}
    \vspace{2mm}
    \label{tab:universal_models}
    \begin{tabular}{lc}
        \toprule
        Universal model & Accuracy (\%) \\
        \midrule
        ResNet-50 (ImageNet) & $74.45 \pm 0.16$ \\
        ViT-S/16 (DINOv3) & $73.99 \pm 0.60$ \\
        ViT-Base (CLIP) & $74.80 \pm 0.30$ \\
        \bottomrule
    \end{tabular}
\end{wraptable}

We also investigate the robustness of GBV to the choice of the universal model $m^u$ for data augmentation. To capture a diverse range of inductive biases and training strategies, we select three distinct architectures from the \texttt{timm} library: ResNet-50 (ImageNet-pretrained)~\citep{he2016deep}, ViT-S/16 (DINOv3)~\citep{simeoni2025dinov3}, and ViT-Base (CLIP)~\citep{radford2021learning}. Following the experiment setup in Section~\ref{sec:exp_aug}, we train the final models on CUB-200-2011 for three runs and report the average performance in Table~\ref{tab:universal_models}. The result shows that GBV remains robust to the choice of $m^u$, consistently outperforming the baselines in Table~\ref{tab:gbv_aug}. Notably, the multimodal pre-training of the CLIP-based model yields a significant improvement, suggesting that cross-modal feature representations are particularly effective for evaluating data augmentation strategies.

\end{document}